\documentclass[sn-mathphys]{sn-jnl}

\usepackage{bbm}
\usepackage{amsmath}
\usepackage{amssymb}
\usepackage{subfig}
\usepackage{float}
\usepackage{graphicx}
\jyear{2021}%

\theoremstyle{thmstyleone}%
\theoremstyle{thmstyletwo}%

\theoremstyle{thmstylethree}%

\begin{document}

\title[Article Title]{Contrast with Major Classifier Vectors for Federated Medical Relation Extraction with Heterogeneous Label Distribution}


\author*[1]{\fnm{Chunhui} \sur{Du}}\email{chunhui18@sjtu.edu.cn}

\author[1]{\fnm{Hao} \sur{He}}\email{hehao@sjtu.edu.cn}

\author[1]{\fnm{Yaohui} \sur{Jin}}\email{jinyh@sjtu.edu.cn}

\affil*[1]{\orgdiv{MoE Key Lab of Artificial Intelligence, AI Institute}, \orgname{Shanghai Jiao Tong University}, \orgaddress{\street{800 Dongchuan RD}, \city{Shanghai}, \postcode{200240}, \state{Shanghai}, \country{China}}}


\abstract{Federated medical relation extraction enables multiple clients to train a deep network collaboratively without sharing their raw medical data. In order to handle the heterogeneous label distribution across clients, most of the existing works only involve enforcing regularization between local and global models during optimization. In this paper, we fully utilize the models of all clients and propose a novel concept of \textit{major classifier vectors}, where a group of class vectors is obtained in an ensemble rather than the weighted average method on the server. The major classifier vectors are then distributed to all clients and the local training of each client is Contrasted with Major Classifier vectors (FedCMC), so the local model is not prone to overfitting to the local label distribution. FedCMC requires only a small amount of additional transfer of classifier parameters without any leakage of raw data, extracted representations, and label distributions. Our extensive experiments show that FedCMC outperforms the other state-of-the-art FL algorithms on three medical relation extraction datasets.}

\keywords{federated learning, relation extraction, contrastive learning, pre-trained language modeling, heterogeneous label distribution}



\maketitle

\section{Introduction}
Neural relation extraction has achieved great success in the past years with the development of deep learning, which usually requires collecting a large amount of text from multiple clients (e.g., hospitals, medical centers) for centralized training. However, this approach may lead to the leakage of patients' privacy, and a secure and widely participated method for data collection is urgently desired. Federated learning (FL) \cite{FedAvg} is a distributed collaborative training paradigm that only exchanges models rather than raw data. Specifically, FL takes multiple rounds of local updates and global aggregations to collaborate with numerous isolated clients. During the local update, a subset of selected clients download the global model from the central server and update it on local private data. The global aggregation is taken by the central server after receiving these local model updates. These two procedures are iterated until convergence.

However, because the data is collected independently by each client, the data distribution across clients is non-independent identical (non-IID). There are many non-IID regimes, such as heterogeneous label distribution, heterogeneous feature distribution, and concept shift across clients\cite{advance}. In this paper, we mainly focus on the common and challenging heterogeneous label distribution, where the distribution of labels varies across clients. As an example, research cases in NIDDK \cite{NIDDK} are primarily about digestive and kidney diseases, while they are about cancer in NIH \cite{NIH}. Heterogeneous label distribution typically severely degrades performance when applying vanilla FL algorithm, FedAvg \cite{FedAvg}. This is because the update by minimizing the local objective function is only based on local data on each client. However, the inconsistent objective functions among clients lead the global model to converge to a stationary point far from global optima achieved by centralized training.

To mitigate inconsistent objective functions among clients, existing methods add various regular terms based on FedAvg to restrict the local models so that they do not deviate from the global model too much \cite{SCAFFOLD}. Nonetheless, a recent empirical study has shown that these modified algorithms do not significantly outperform vanilla FedAvg on most of the datasets \cite{experimental}. We believe there are two limitations that hinder performance improvements. One is that the regular terms exist only between the local and global models, while other clients' models are ignored. This may be due to communication costs and privacy concerns, but greater performance improvements may be obtained if other clients' models can be leveraged efficiently and securely. The other is that these methods may be applicable to general non-IID cases without being specifically designed for heterogeneous label distribution. For the typical relation extraction network, the features of two entities are extracted and concatenated together by the feature extractor, and then the relation class is determined by calculating the similarity with a set of class vectors in the classifier\cite{match}. In fact, the classifier is particularly important for heterogeneous label distribution, as shown in previous studies \cite{FedRS, nofear}. We propose the novel concept of \textit{major classifier vectors}, where each class vector is picked from the best one among all clients. Generally, the more samples of a particular class for each client, the better the class vector. For example, suppose there are two relation classes and two clients, where client A contains mainly class 1 samples and client B contains mainly class 2 samples. We obtain the ensemble major classifier vectors by combining client A's class 1 vector and client B's class 2 vector rather than the weighted average method as shown in FedAvg\cite{FedAvg}. The introduction of major classifier vectors fully leverages the models of multiple clients and requires minimal communication and privacy costs.

Unfortunately, the label distribution of each client is agnostic in FL, so we cannot obtain the major classifier vectors directly based on the number of samples. Inspired by \cite{FedRS}, we find that the model is hard to distinguish minor classes (i.e., the classes with fewer samples), whose extracted representations are easily collapsed and mixed. On the contrary, the model has higher accuracy for major classes (i.e., the classes with more samples), and the corresponding extracted representations are far away from that of minor classes. Accordingly, the classifier vector of the major class usually also has low similarity to other classifier vectors. Based on this idea, major classifier vectors can be obtained in the server based on inter-client classifier similarity under agnostic label distribution. The major classifier vectors are then distributed to all clients along with the aggregated global model. Then the local training of each client is Contrasted with Major Classifier vectors (FedCMC), so the local model is not prone to overfitting to the local label distribution.

Our main contributions include the following:

1. We present the problem of federated medical relation extraction and focus on the challenge of heterogeneous label distribution. 

2. To alleviate this challenge, we propose FedCMC, which obtains major classifier vectors in the server based on inter-client similarity and constrains local training with it in the client. FedCMC requires only a small amount of additional transfer of classifier parameters without any leakage of raw data, extracted representations, and label distributions.

3. Through comprehensive experiments, we demonstrate that FedCMC can drastically benefit the performance and convergence speed of FL models.

\section{Related Work}
\subsection{Medical Relation Extraction}
Reading text to identify and extract relations between entities has been a long-standing goal in natural language processing \cite{RE}. A variety of neural network approaches have been applied to relational extraction, including CNN\cite{CNN}, RNN\cite{RNN}, GCN\cite{GCN}. and recent pre-trained language modeling (PLM) \cite{match}. With significant growth in the medical literature, the application of relation extraction models to aid in the analysis of electronic medical records or biomedical reports has great potential \cite{medical1, medical2}.

\subsection{Federated Learning}
Medical data often sits in geographically distributed data silos, and privacy concerns restrict the use of this data. The traditional centralized training paradigm faces significant challenges. FL is an emerging learning paradigm seeking to address the problem of data governance and privacy by training algorithms collaboratively without exchanging the data itself \cite{advance}. FedED\cite{FedED} and Lazy MIL\cite{distant} have studied supervised and distant federated relation extraction, respectively. The vanilla FL algorithm, FedAvg, periodically aggregates the local models in the server and updates the local model with its individual data. FedProx\cite{FedProx} adds a proximal term to the local subproblem to restrict the local update closer to the global model. SCAFFOLD\cite{SCAFFOLD} uses a variance reduction technique to correct the drifted local update. FedDyn \cite{FedDyn} modifies the objective of the client with linear and quadratic penalty terms to align global and local objectives. The above approach can be applied to any non-IID data in FL but generally has only a small boost for heterogeneous label distributions. FedRS\cite{FedRS} considers the classification layer to be more vulnerable to such distributions than the feature extraction layer and therefore limits the degree of updates of the classifier by modifying the standard softmax. Similarly, the more recent FedLC\cite{FedLC} also modifies the softmax by reducing the logits (i.e., the output of the last classification layer and the input to softmax) of the major class while increasing the logits of the minor class. However, these methods ignore the interaction with other clients and only perform well when the local label distribution is imbalanced while the global label distribution is relatively balanced \cite{TALT}.

\subsection{Contrastive Learning}
Contrastive learning has shown great promise in self-supervised representation learning, which reduces the distance between the representations of different augments of the same sample (i.e., \textit{positive pairs}), and increases the distance between the representations of augments of different samples (i.e., \textit{negative pairs}). There are two typical kinds of contrastive learning, one is within-batch contrast which requires large batch size (e.g., SimCLR\cite{SimCLR}), and the other is cross-batch contrast, where momentum mechanism is required (e.g., MOCO\cite{MOCO}). Besides, the contrast loss is usually optimized together with the cross-entropy loss for supervised learning, which usually yields more robust representations\cite{SCL}. There are also several works considering contrastive learning in FL. For medical image segmentation, FedDG\cite{FedDG} considers the same boundary-related or background-related representations as positive pairs among clients and otherwise negative pairs. However, this method is specialized for the image segmentation task, and sharing the magnitude spectrum of all clients carries the risk of privacy compromise. MOON\cite{MOON} tries to decrease the distance between the representation learned by the local model and the representation learned by the global model and increase the distance between the representation learned by the local model and the representation learned by the previous local model. However, contrasting with the representation extracted by the global model only has minor impacts, especially for heterogeneous label distribution\cite{experimental}.

\section{Method}
\subsection{Task Definition}
\label{section1}
Given a sentence $s=\{w_0,w_1,...,w_n\}$ with an entity pair $e_1=\{w_i,...,w_j\}$ and $e_2=\{w_k,...,w_l\}$, relation extraction task aims to predict the relation type between $e_1$ and $e_2$, where $C$ denotes the number of relation types. Following recent studies\cite{match, type}, we utilize the PLM with parameters $\Theta^E$ as the backbone encoder. And we construct the input sequence $\hat{s}=\{[CLS],w_0,...,<e1>,w_i,...,w_j,</e1>,...,<e2>,w_k,...,w_l,</e2>,...,w_n,[SEP]\}$. 

Given the prepared sequence $\hat{s}$ as input, the output of the PLM encoder is expressed as $H\in\mathbbm{R}^{m\times d}$, where $m$ is the input sequence length, and $d$ is the output dimension of the encoder. We obtain entity representations $h_{e_1}=\sum([h_i,..h_j])\in \mathbbm{R}^d$ and $h_{e_2}=\sum([h_k,...,h_l])\in \mathbbm{R}^d$ by summing the outputs of word pieces in each entity. And the relation representation $h$ if obtained by concatenating two entity representations
\begin{equation}
    h=h_{e_1}\oplus h_{e_2}\in \mathbbm{R}^{2d}
\end{equation}

Let $\Theta^L\in \mathbbm{R}^{C,2d}$ denote the trainable parameters of the classifier, the probability score on label $y$ is
\begin{equation}
p(y\vert\hat{s};\Theta^E,\Theta^L)=\frac{exp(\Theta^L_y\cdot h)}{\sum_{c=1}^{C}exp(\Theta^L_c\cdot h)}
\end{equation}
where $\Theta^L_c$ is the classifier vector of class $c$.

Suppose there are $N$ samples. We denote $\hat{s}_i$, $y_i$, and $h_i$ as the input, label, and extracted relation representation of the $i$-th sample. The cross-entropy loss is calculated as
\begin{equation}
    L_{ce}=-\frac{1}{N}\sum_{i=1}^{N}\sum_{c=1}^{C}\mathbf{I}_{c=y_i}\log p_{i,c}
\label{ce loss}
\end{equation}

where $p_{i,c}:=p(c\vert\hat{s}_i;\Theta^E,\Theta^L)$. The gradient of $\Theta^L_c$ is

\begin{equation}
\frac{\partial L_{ce}}{\partial \Theta^L_c} = -\sum_{i=1}^{N} (\mathbf{I}_{c=y_i}-p_{i,c})\cdot h_i
\end{equation}

We use gradient descent with learning rate $\eta$ to update $\Theta_c^L$ and decompose this update into the pulling and pushing forces following FedRS\cite{FedRS}:
\begin{equation}
\Theta^L_c\leftarrow \Theta^L_c+ \eta \sum_{i=1,y=c}^{N}(1-p_{i,c})\cdot h_i - \eta \sum_{i=1,y\neq c}^{N}(1-p_{i,c})\cdot h_i
\end{equation}

The classifier vector $\Theta^L_c$ is pulled close to extracted relation representations with label $c$ while pushed away from those with other labels. Label imbalance of each client is very common for heterogeneous label distribution. We refer to the major class as the class with a large number of samples and the minor class as the class with a small number of samples. Thus, the classifier vector of the major class is mainly affected by pushing from samples of the major class, while the classifier vector of the minor class is mainly affected by pulling away from samples of the major class. This is reflected by the fact that the classifier vector of the major class has low similarity to other minor class vectors, while the similarity between classifier vectors of the minor class.

\subsection{Major Classifier Vectors}

FL has already shown great promise for cross-institutional healthcare research with a privacy-preserving scheme \cite{npj, nature}. Suppose there are $K$ medical clients. Let $N^c_k$ denote the number of samples with label $c$ in client $k$. The number of samples in client $k$ is $N_k=\sum_{c=1}^{C}N_k^c$ and the number of samples with label $c$ of all clients is $N^c=\sum_{k=1}^{K}N^c_k$. Each client $k$ has a private relation extraction dataset $\mathcal{D}_k=\{\{s_{k,1},y_{k,1}\},..,\{s_{k,N_k},y_{k,N_k}\}\}$.

In each communication round $t$, the global model parameters, including PLM encoder parameters $\Theta^{E,t}$ and classifier parameters $\Theta^{L,t}$, are distributed to each client $k\in[1,...,K]$ as $\Theta^{L,t}_k$ and $\Theta^{L,t}_k$. Then $\Theta^{E,t}_k$ and $\Theta^{L,t}_k$ are updated in multiple local epochs

\begin{equation}
    \Theta^{E,t}_k=\Theta^{E,t}_k-\eta \frac{\partial L_{ce}}{\partial \Theta^{E,t}_k}
\end{equation}
\begin{equation}
    \Theta^{L,t}_k=\Theta^{L,t}_k-\eta \frac{\partial L_{ce}}{\partial \Theta^{L,t}_k}
\end{equation}

The aggregation of the global encoder and classifier in round $t$ is the same as FedAvg
\begin{equation}
\Theta^{E,t+1}=\sum_{k=1}^{K}p_k\Theta^{E,t}_k
\label{global1}
\end{equation}

\begin{equation}
\Theta^{L,t+1}=\sum_{k=1}^{K}p_k\Theta^{L,t}_k
\label{global2}
\end{equation}
where the aggregation weight $p_k$ is determined by the specific FL algorithm. In most cases, $p_k=\frac{N_k}{\sum_{k=1}^{K}N_k}$.

As discussed in \ref{section1}, the similarity between classifier vectors can measure the number of samples to a certain extent. Therefore we introduce the \textit{local average similarity} and the \textit{global average similarity} as follows.

\textbf{Definition 1}. For each client $k\in[1,..K]$, \textit{local average similarity} of classifier vector $\Theta_k^{L,c}$  with class $c$ to the other classifier vectors $\{\Theta_k^{L,i},i\in[1,...,C],i\neq c\}$ is defined as
\begin{equation}
    d_{k,c}^t=\frac{1}{C - 1}\sum_{i=1,i\neq c}^{C} \Theta_{k,c}^{L}\odot \Theta_{k,c}^{L}
\end{equation}
where $\odot$ denote the cosine similarity: $a\odot b=\frac{<a,b>}{\Vert a\Vert_2\Vert b\Vert_2}$.

\textbf{Definition 2} According to the aggregation of $\Theta_k^L$ in Eq. (\ref{global1}), \textit{global average similarity} with class $c$ is defined as
\begin{equation}
    d^t_c=\frac{1}{K}\sum_{k=1}^{K} \frac{N_k}{N} d^t_{k,c}
\end{equation}

\begin{figure*}[htbp]
\centering
	\subfloat{\includegraphics[width = 0.48\textwidth]{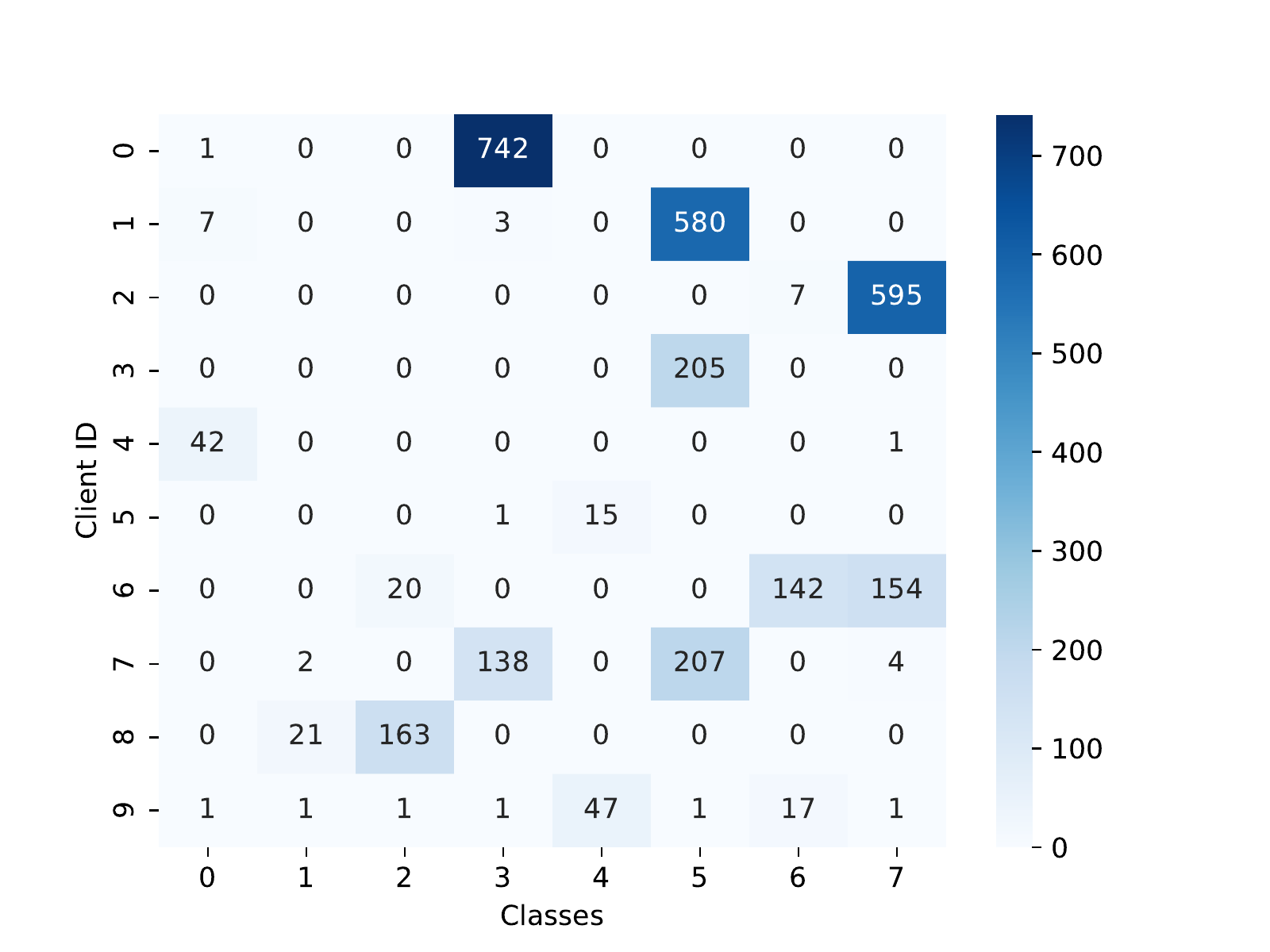}}
	\subfloat{\includegraphics[width = 0.48\textwidth]{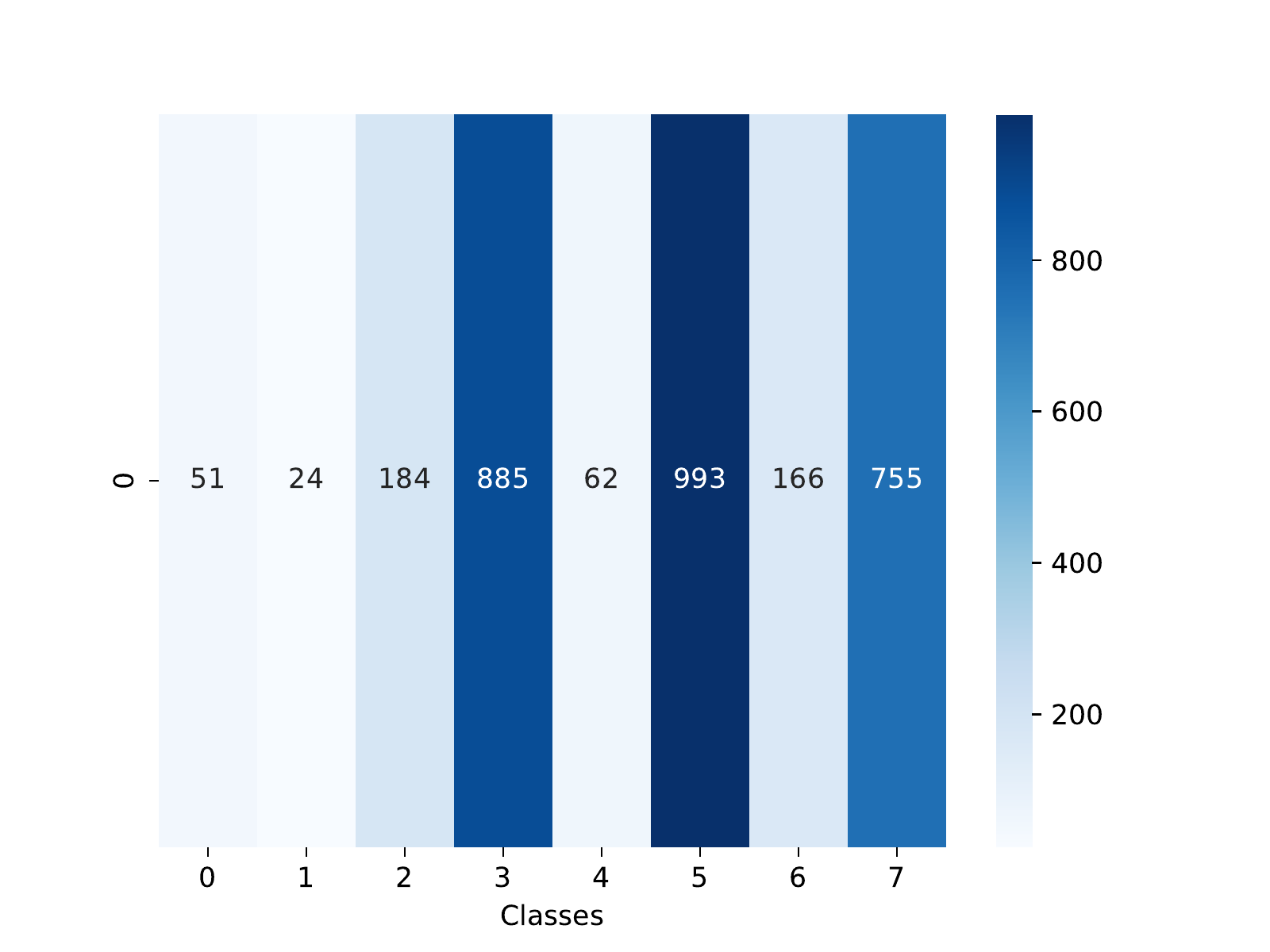}}
	\\
	\subfloat{\includegraphics[width = 0.48\textwidth]{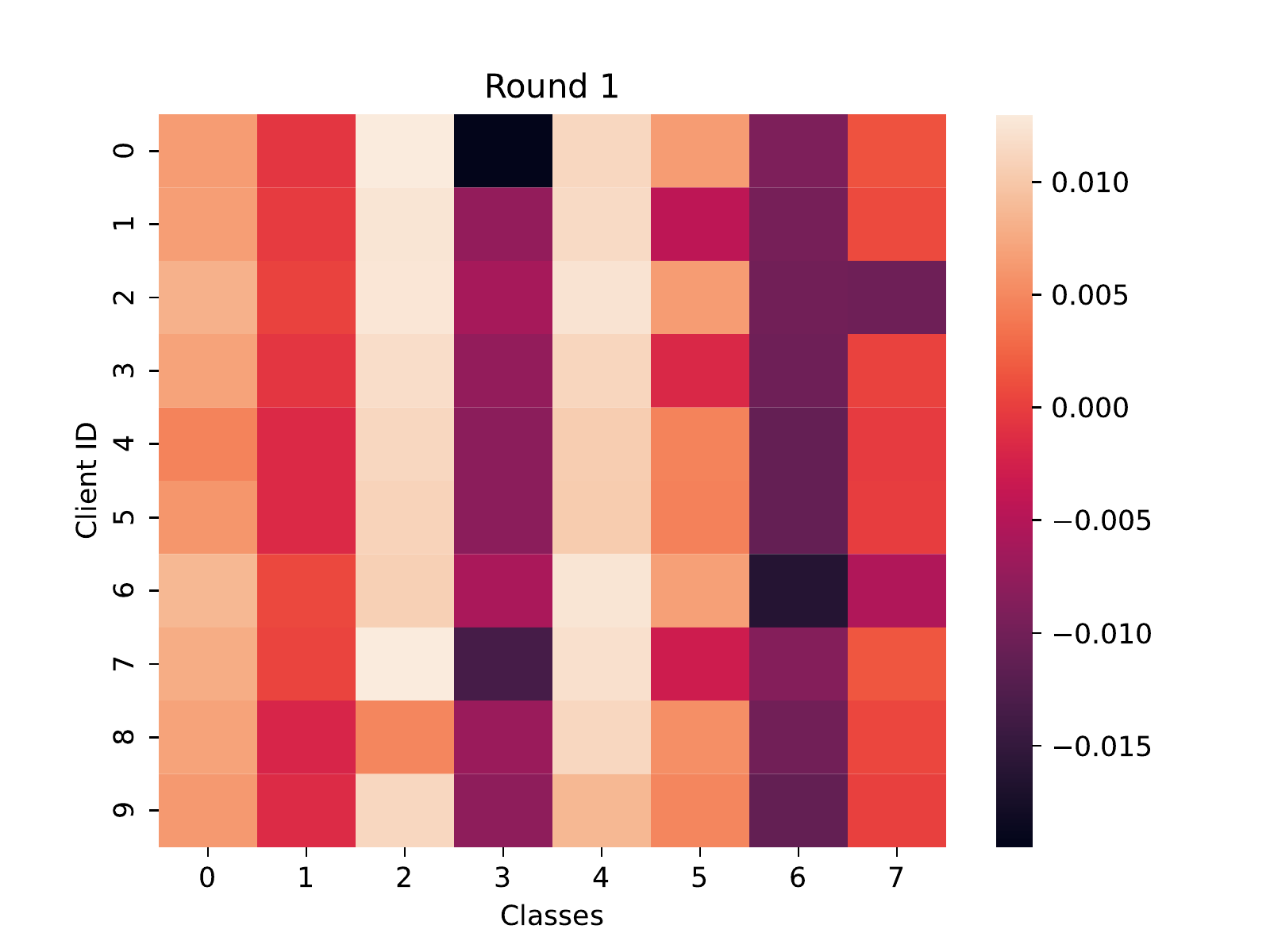}}
	\subfloat{\includegraphics[width = 0.48\textwidth]{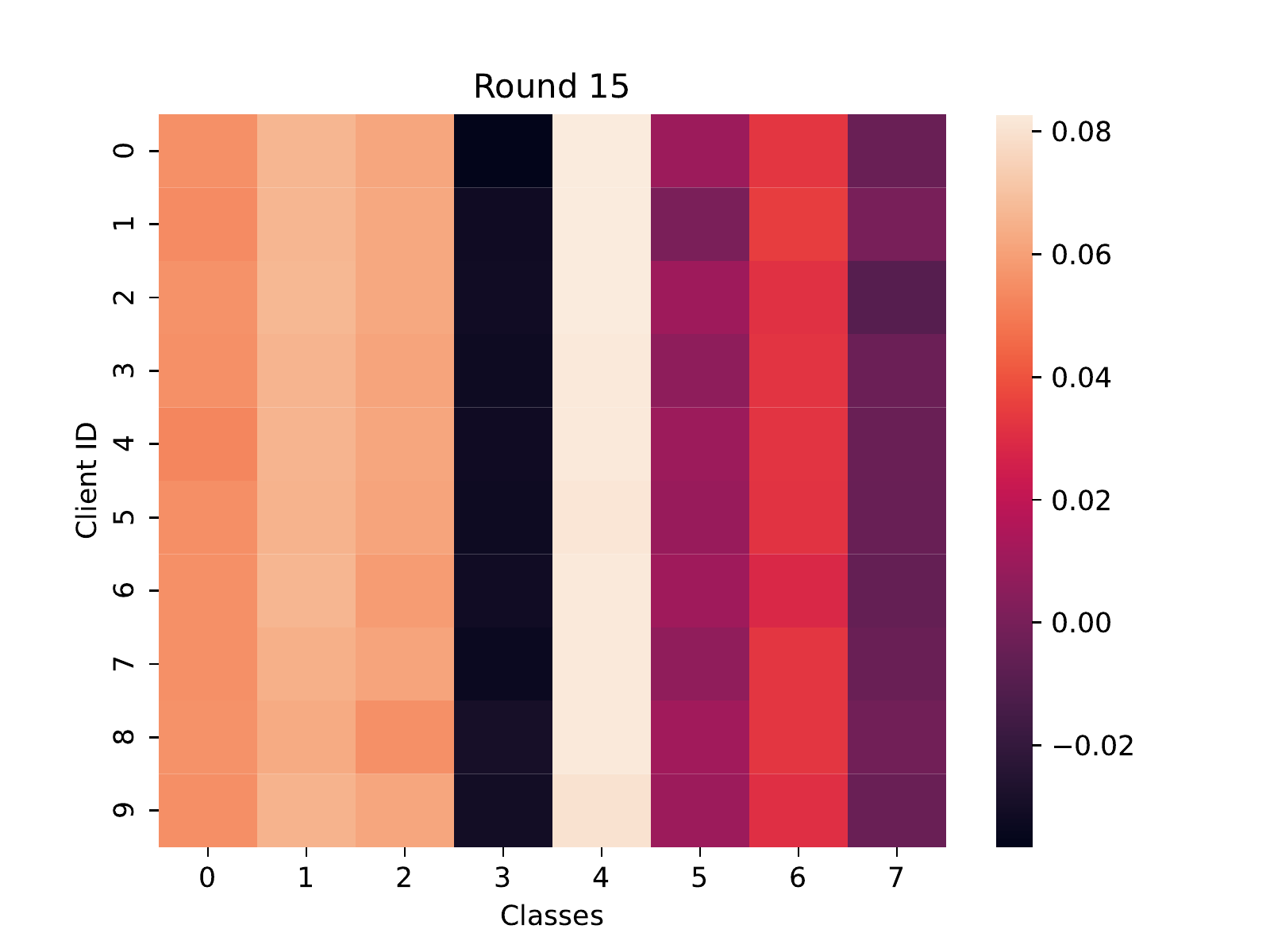}}
\caption{Local label distributions (a) and global distribution (b) when $\alpha=0.05$ for 2010 i2b2/VA challenge. local average similarity $d_{k,c}^t$ of client $k$ and class $c$ when communication round $t=1$ (c) and $t=15$ (d).}
\label{sim}
\end{figure*}

To vividly understand how $d^t_{k,c}$ and $d^t_{k,c}$ vary with communication rounds, we implemented an empirical study in the 2010 i2b2/VA challenge dataset with heterogeneous label distributions. Specifically, we set $K=10$ and partition the data with 8 relation classes according to the Dirichlet distribution with the concentration parameter $\alpha=0.05$. We can obtain the following three observations from Fig. \ref{sim}.

\textbf{Observation 1}. In the first few communication rounds, $\forall k\in[1,...,K],  \forall c1,c2\in[1,...,C]$, when $N^{c1}_{k}$ is greater than $N^{c2}_{k}$, $d_{k,c1}^t$ is usually less than $d_{k,c2}^{t}$. 

\textbf{Observation 2}. As the increases in communication rounds, $\forall k\in[1,...,K]$, $d_{k,c}^t$ is closer to $d_c^t$. $\forall c1,c2\in[1,...,C]$, when $N^{c1}$ is greater than $N^{c2}$, $d^{t}_{c1}$ is usually less than $d^{t}_{c2}$. 

\textbf{Observation 3}. In all communication rounds, $\forall c\in[1,...,C]$, $\forall k1,k2\in[1,...,K]$, when $N^c_{k1}$ is greater than $N^c_{k2}$, $d^t_{c, k1}$ is usually less than $d^t_{c, k2}$.

Observations 1 and 2 illustrate the association of $d^t_{k,c}$ with $N_K^c$ across different classes for each client, i.e., the row of the matrix in Fig. \ref{sim} (a). In the initial stage of training, the classifier vectors are mainly influenced by the local label distribution. As the parameters are aggregated over multiple rounds on the server, the classifier vectors are mainly determined by the global label distribution. 

Although observations 1 and 2 show that the distribution of $\{d^t_{k,c}, \forall c\in[1,..., C]\}$ crossing classes varies greatly with the number of communication rounds, observation 3 illustrates that the distribution of $\{d^t_{k,c}, \forall k\in[1,..., K]\}$ crossing clients varies very little, i.e., the column of the matrix in Fig. \ref{sim} (a). Specifically, if client $k$ contains a majority of samples with class $c$ across all clients, then $d^t_{k,c}$ is minimal among clients in all communication rounds. Therefore, we propose the \textit{major classifier vector} $\hat{\Theta}^{L,t}_{c}$ with class $c$ and \textit{major classifier vectors} $\hat{\Theta}^{L,t}$ next.

\textbf{Definition 3} $\forall t\in[1,..,T]$, $\forall c\in[1,..,C]$, given the set of $\{d^t_{1,c},...,d^t_{K,c}\}$, the \textit{major classifier vector} with class $c$ is defined as
\begin{equation}
\label{min}
    \hat{\Theta}^{L,t}_{c}=\arg\min_{\Theta^{L,t}_{1,c},...,\Theta^{L,t}_{K,c}} \{d^t_{1,c},...,d^t_{K,c}\}
\end{equation}

\textbf{Definition 4} The set of \textit{major classifier vectors} is defined as
\begin{equation}
    \hat{\Theta}^{L,t}=\{\hat{\Theta}^{L,t}_1;...;\hat{\Theta}^{L,t}_C\}
\label{vectors}
\end{equation}

Compared to $\Theta^{L,t}$, $\hat{\Theta}^{L,t}$ is obtained by major classifier vectors ensemble and is not affected by the aggregation weights $p_k$. Thus, in addition to $\Theta^{E,t}$ and $\Theta^{L,t}$, we also distribute $\hat{\Theta}^{L,t}$ to all clients.

\subsection{Framework}

\begin{figure}[t]
\centering
\includegraphics[width=0.95\textwidth]{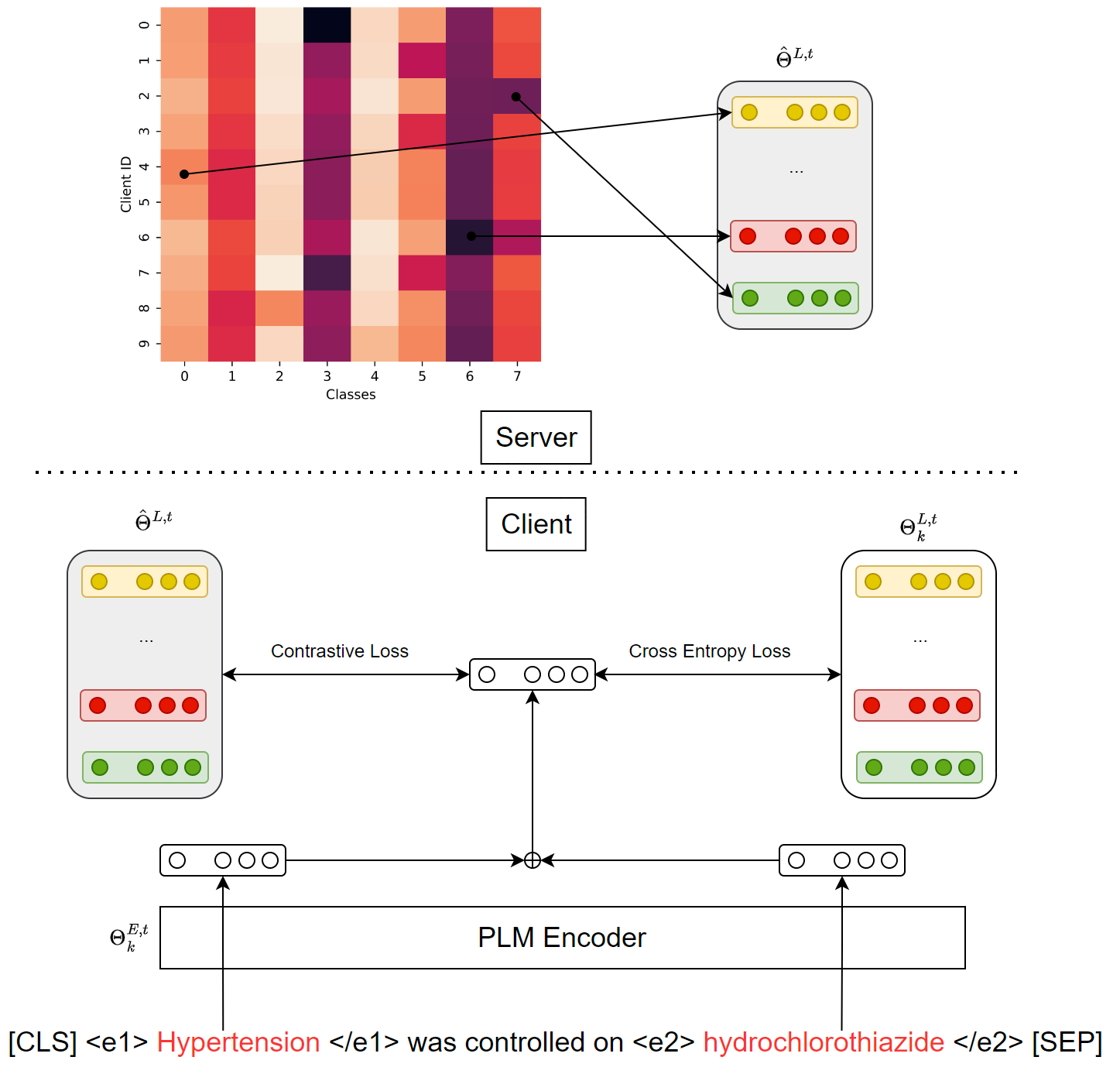}
\caption{The architecture of FedCMC. The major classifier vectors $\hat{\Theta}^{L,t}$ are computed in the server and distributed to all clients.}
\label{architecture}
\end{figure}

\begin{algorithm}[t]
\renewcommand{\algorithmicrequire}{\textbf{Input:}}
\renewcommand{\algorithmicensure}{\textbf{Output:}}
\footnotesize
\caption{Framework}
\begin{algorithmic}[1]
\Require $T$, $K$, $E$, $\eta$;
\Ensure $\Theta^{E,T}$, $\Theta^{L,T}$;

\State initialize $\Theta^{E,0}$, $\Theta^{L,0}$
\State initialize $\hat{\Theta}^{L,0}$=$\Theta^{L,0}$
\For{$t=0,1,...,T-1$}
    \For{$k=1,2,...,N$ in parallel}
        \State $\Theta^{E,t}_k$, $\Theta^{L,t}_k$ $\Leftarrow$ 
        LocalTraining($\Theta^{E,t}$, $\Theta^{L,t}$, $\hat{\Theta}^{L,t}$);
    \EndFor
    \State compute $\Theta^{E,t+1}_k$, $\Theta^{L,t+1}_k$ 
    according to Eq. (\ref{global1}) and Eq. (\ref{global2})
    \State compute $\Theta^{\hat{L},t+1}_k$ according to Eq. (\ref{vectors})
\EndFor
\State return $\Theta^{E,T}$, $\Theta^{L,T}$;
\State
\State \textbf{LocalTraining}:
\For{epoch i=1,2,...,E}
    \State compute $L_{sup}$ according to Eq. (\ref{ce loss});
    \State compute $L_{con}$ according to Eq. (\ref{con loss});
    \State compute $L$ according to Eq, (\ref{sum})
    \State update $\Theta^{E,t}_k$ and $\Theta^{L,t}_k$ according to Eq. (\ref{update1}) and Eq. (\ref{update2});
\EndFor
\end{algorithmic}
\label{algorithm}
\end{algorithm}

For each client $k\in[1,...,K]$, we define the contrastive loss between extracted features $h_i$ and major classifier vectors $\hat{\theta}^L$ for each sample $i$ with label $y$ as follows 
\begin{equation}
L_{con}=\sum_{i=1}^{N_k}-log\frac{exp(h_i\cdot \hat{\Theta}^{L}_{y})}{\sum_{c=1}^{C} exp(h_i\cdot  \hat{\Theta}^{L}_c)}   
\label{con loss}
\end{equation}

The final loss is computed by
\begin{equation}
    L=L_{ce}(\Theta^E, \Theta^L) + \mu L_{con}(\Theta^E, \hat{\Theta}^L)
\label{sum}
\end{equation}
where $\mu$ is a hyper-parameter to control the weight of contrastive loss.

The PLM encoder parameters $\Theta^{E,t}_k$ and classifier parameters $\Theta^{L,t}_k$ are updated as follows

\begin{equation}
    \Theta^{E,t+1}_k=
    \Theta^{E,t}_k-\eta \frac{\partial L}{\partial \Theta^{E,t}_k}=
    \Theta^{E,t}_k-\eta(\frac{\partial L_{ce}}{\partial \Theta^{E,t}_k}+\mu\frac{\partial L_{con}}{\partial \Theta^{E,t}_k})
\label{update1}
\end{equation}

\begin{equation}
    \Theta^{L,t+1}_k=
    \Theta^{L,t}_k-\eta \frac{\partial L}{\partial \Theta^{L,t}_k}=
    \Theta^{L,t}_k-\eta \frac{\partial L_{ce}}{\partial \Theta^{L,t}_k}
\label{update2}
\end{equation}

 The overall framework is shown in Algorithm \ref{algorithm}, and the architecture of FedCMC is shown in Fig. \ref{architecture}. In each communication round, each client uses stochastic gradient descent to update the global model $\Theta_k^{E, t}$ and $\Theta_k^{L,t}$ with its local data. A significant difference from FedAvg is that the PLM encoder $\Theta_k^{E, t}$ is supervised by the ensemble major classifier vectors $\hat{\Theta}^{L, t}$ so that the learned representations are more unbiased. After receiving the local model parameters from all clients, the server aggregates them into the global model parameters $\Theta_k^{E,t+1}$ and $\Theta_k^{L,t+1}$ for the next round. In addition, the major classifier vectors $\hat{\Theta}^{L, t+1}$ are obtained based on the similarity calculation. 

\section{Experiments}
\subsection{Experiment Setup}

\begin{table}
\footnotesize
\caption{Statistics of datasets}
\label{sta}
\tabcolsep 20pt
\begin{tabular*}{\textwidth}{ccccc}
\toprule
Dataset & \# train & \# test & \# relations\\
\hline
2010 i2b2/VA challenge & 3120 & 6293 & 8 & \\
CPR & 10k & 2k & 5 & \\
PGR & 3436 & 860 & 2\\
\bottomrule
\end{tabular*}
\end{table}

\begin{itemize}
    \item \textit{Datasets} We evaluate FedCMC on three well-known medical relation extraction datasets, including the 2010 i2b2/VA challenge dataset \cite{2010}, BioCreative VI: Chemical-protein interaction (CPR)\cite{CPR}, and Phenotype-Gene Relations (PGR) corpus \cite{PGR}. The 2010 i2b2/VA challenge dataset collects clinical records and is used to classify relations between medical problems, tests, and treatments. There are 9413 sentences and 8 types of relations: TrIP, TrWP, TrCP, TrAP, TrNAP, TeRP, TeCP, and PIP. For example, \textit{the tumor was growing despite the available chemotherapeutic regimen} implies the relation TrWP, i.e., treatment (chemotherapeutic) worsens medical problem (tumor). It is worth noting that the test set has 6293 samples, more than the 3120 samples in the training set as released in \cite{2010}. CPR provides 10307 sentences about what a chemical does to a gene/protein. There are 5 types of relations: CPR:3, CPR:4, CPR:5, CPR:6, and CPR:9. PGR is a widely used biomedical corpus, which contains 4296 sentences to determine if a relation between human phenotype and gene exists. In addition, all validation sets were randomly sampled by 10\% from the training set. Table \ref{sta} presents the statistical information of the three datasets. Like previous studies \cite{Bayesian, MOON}, we use Dirichlet distribution to generate the heterogeneous data partition among clients. Specifically, we sample $p_k\sim Dir_K(\alpha)$ and allocate a $p_{k,c}$ proportion of the instances of class $c$ to client $k$, where $Dir_K(\alpha)$ is the Dirichlet distribution with a concentration parameter $\alpha$. The smaller the $\alpha$, the greater the heterogeneity of the label distribution. We set the default number of clients $K$ to $10$, the concentration parameter $\alpha=0.1$, and $\alpha=0.01$. With the above partitioning strategy, different clients have different data samples for the same class.
    \item \textit{Baselines} We compare our algorithm FedCMC with two well-known FL algorithms FedAvg \cite{FedAvg}, MOON \cite{MOON}. Besides, two FL algorithms proposed specifically to address the heterogeneous label distribution, FedRS \cite{FedRS} and FedLC \cite{FedLC}, are also considered.
    \item \textit{Implementation Details} Considering the computation and communication of multiple rounds for multiple clients in FL, we use DistilBERT \cite{distilbert} in our experiments as the trade-off between performance and cost. DistilBERT is a distilled version of the BERT model and has a 60\% faster inference speed and 40\% smaller model size. To conduct a fair comparison, we list all federated methods hyper-parameters as follows. The learning rate is set to 5e-5, the batch size is fixed to 8, and the number of local epochs is set to 1. All experiments are run with NVIDIA GeForce RTX 3090 and repeated 3 times with different random seeds. 
\end{itemize}
\subsection{Results Comparison}
We run a large number of experiments to demonstrate the superiority of our algorithm on performance and convergence speed for different label heterogeneity. In addition, our algorithm is robust under different settings of FL, including the number of local epochs and the number of clients. 

\begin{table}
\footnotesize
\caption{Results on 2010 i2b2/VA challenge, CPR and PGR for different label heterogeneity.}
\label{table}
\tabcolsep 22pt
\begin{tabular*}{\textwidth}{cccc}
\toprule
  Algorithm & 2010 i2b2/VA challenge & CPR & PGR\\\hline
  centralized & 63.58 & 55.7 &73.30 \\\hline
  \multicolumn{4}{c}{$\alpha=0.5$, $K=10$}\\\hline
  FedAvg & 52.34 & 45.91  & 70.26\\
  MOON & 53.91 & 44.67 & 71.45\\
  FedRS & 55.79 & 49.23 & 72.10\\
  FedLC & 56.02 & 48.20 & 72.06\\
  FedCMC & \textbf{61.42} & \textbf{53.72} & \textbf{73.38}\\\hline
  \multicolumn{4}{c}{$\alpha=0.05$, $K=10$}\\\hline
  FedAvg & 47.56 & 42.82 & 68.36\\
  MOON & 47.92 & 43.06 & 68.02 \\
  FedRS & 51.98 & 45.63 &70.49 \\
  FedLC & 54.20 & 46.91 & 71.82 \\
  FedCMC & \textbf{60.55} & \textbf{51.21} & \textbf{72.40} \\\hline
  \multicolumn{4}{c}{$\alpha=0.05$, $K=20$}\\\hline
  FedAvg & 45.43 & 40.67 & 67.26\\
  FedCMC & \textbf{59.56} & \textbf{51.05} & \textbf{71.67} \\\hline
  \multicolumn{4}{c}{$\alpha=0.05$, $K=50$}\\\hline
  FedAvg & 42.48 & 38.08 & 63.36\\
  FedCMC & \textbf{58.82} & \textbf{50.28} & \textbf{71.56} \\

\bottomrule
\end{tabular*}
\end{table}

\begin{figure}[htbp]
\centering
\subfloat{\includegraphics[width = 0.8\textwidth]{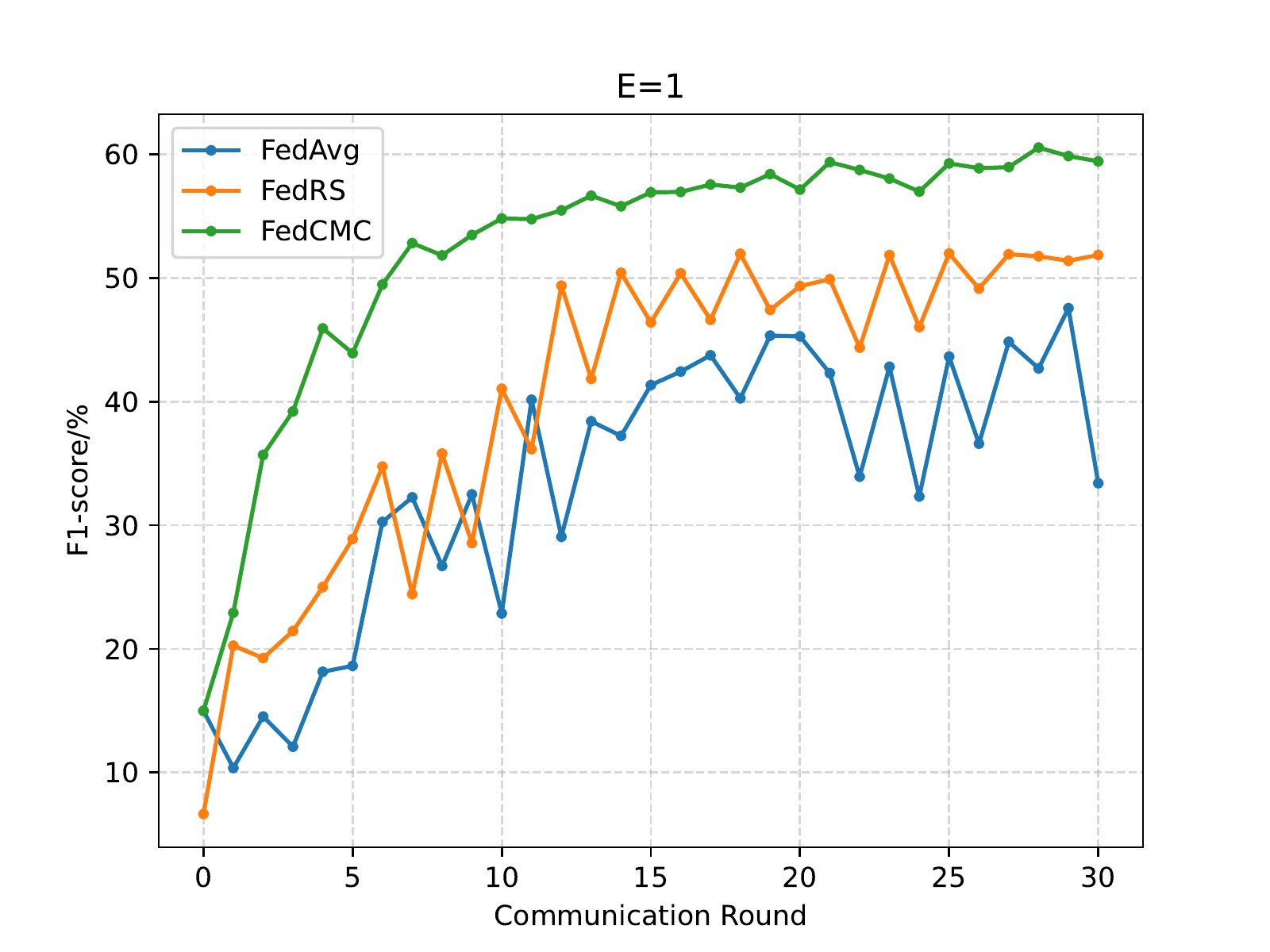}}
\\
\subfloat{\includegraphics[width = 0.8\textwidth]{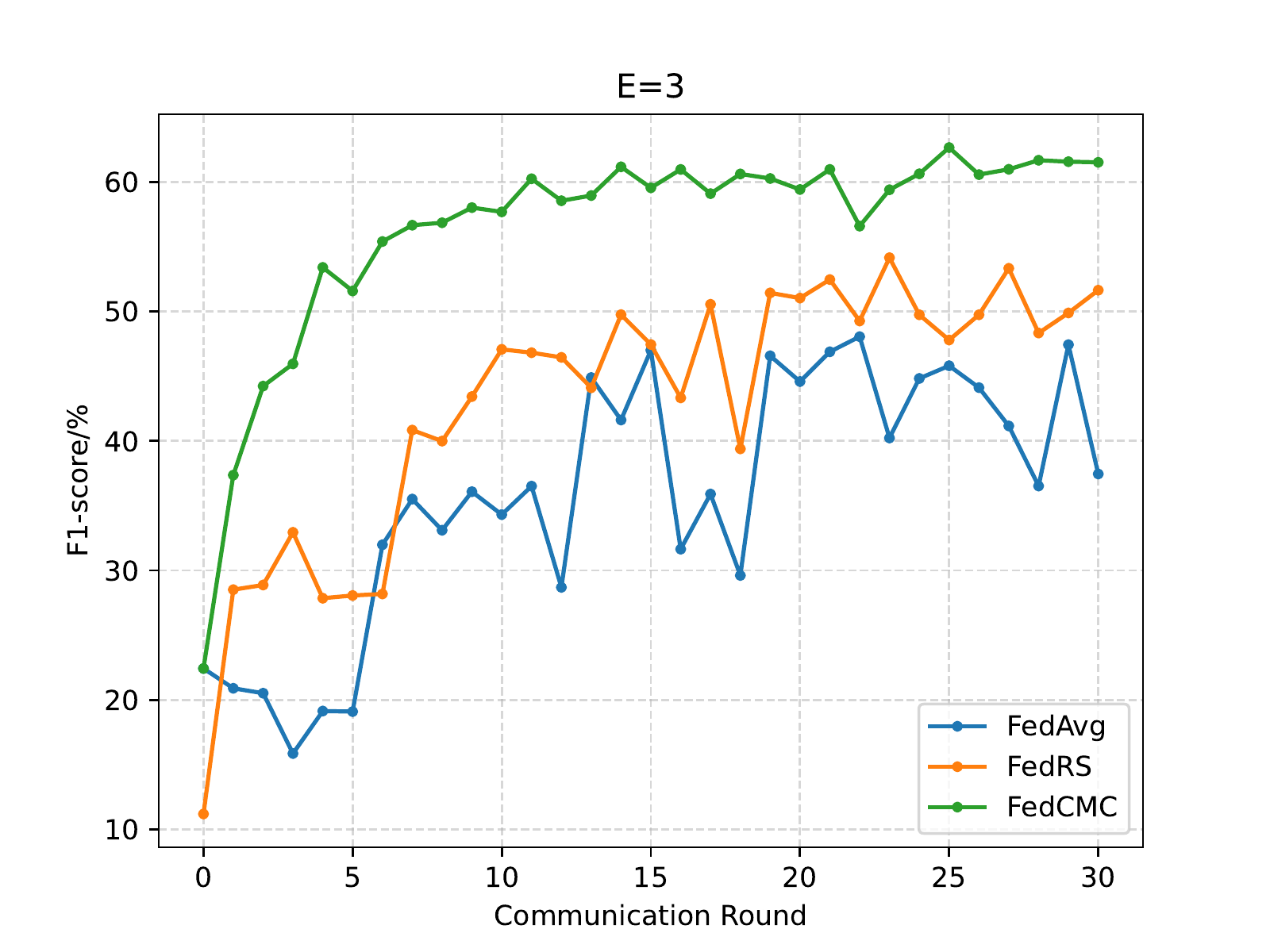}}
\caption{Convergence plots for FedCMC and other baselines when local epochs $E=1$ (a) and $E=3$ (b) when $\alpha=0.05$ for 2010 i2b2/VA challenge dataset.}
\label{train}
\end{figure}

As shown in Table \ref{table}, the algorithms FedRS and FedLC proposed specifically for heterogeneous label distribution is significantly better than FedAvg and MOON. However, our algorithm achieves the best F1-score on all three datasets for both high ($\alpha=0.05$) and low ($\alpha=0.5$) degrees of heterogeneity. Compared with FedAvg, when $\alpha=0.05$, the F1-score of our algorithm is improved by 12.99\%, 8.39\%, and 4.04\% for the 2010 i2b2/VA challenge, CPR, and PGR, respectively. When $\alpha=0.5$, 9.08\%, 7.81\% and 2.12\% improvements are achieved. It can be seen that FedAvg has a significant performance degradation at high heterogeneity($\alpha=0.05$), while our FedCMC algorithm has only a small degradation. Moreover, FedAvg performs worse on the dataset 2010 i2b2/VA challenge with 8 relation types compared to the binary PGR dataset, which is consistent with previous studies\cite{FedRS}. In contrast, our FedCMC is robust on all datasets.

In addition to better performance, our algorithm has a faster convergence speed compared to the mentioned baselines, as shown in Fig. \ref{train} (a). For example, for the target 45\% F1-score, our algorithm requires only 5 communication rounds to achieve the 45\% score when $\alpha=0.05$, while FedAvg and FedRS require 20 and 15, respectively.

We also study the effect of local epochs on the performance, as shown in Fig. \ref{train} (b). As the number of local epochs increases, the convergence speed of FedAvg and FedRS is not significantly faster except for our algorithm. And the variance of the F1-score of FedAvg is significantly larger as the inconsistency between local and global optimal points becomes larger, while our algorithm is relatively robust.

We further analyze how the generalization performance of our algorithm will be affected by different numbers of clients in FL. We conduct experiments to analyze the effectiveness of our algorithm when increasing the number of clients. We report the F1-score in Table with 20 and 50 clients for the 2010 i2b2/VA challenge dataset in Table. \ref{table}. Our algorithm achieves the best performance consistently. Moreover, the performance gap between our algorithm and other baselines increases when the client size increases from 10 to 50.

\subsection{Ablation Analysis}
In this section, we analyze how learning by contrastive learning with major classifier vectors can reduce objective inconsistency across clients and thus improve performance.

The key to our algorithm is to pick out \textit{major classifier vectors} according to Eq. (\ref{min}) and to perform contrastive learning based on them according to Eq. (\ref{con loss}). To evaluate the impact of different classifier vectors on the results, we provide two alternative classifier vectors. One is \textit{random classifier vectors} and the other is \textit{minor classifier vectors}, i.e., the opposite of Eq. (\ref{min}):
\begin{equation}
\label{max}
    \hat{\Theta}^{L,t}_{c}=\arg\max_{\Theta^{L,t}_{1,c},...,\Theta^{L,t}_{K,c}} \{d^t_{1,c},...,d^t_{K,c}\}
\end{equation}

\begin{table}
\footnotesize
\caption{Results on unbalanced and balanced CPR.}
\label{balance}
\tabcolsep 30pt
\begin{tabular*}{\textwidth}{ccccc}
\toprule
F1-score & FedRS & FedLC & FedCMC\\
\hline
unbalanced & 31.2 & 32.7 & 40.5 & \\
balanced & 42.9 & 41.8 & 43.2 & \\
\bottomrule
\end{tabular*}
\end{table}

\begin{figure}[htbp]
\centering
\includegraphics[width = 0.9\textwidth]{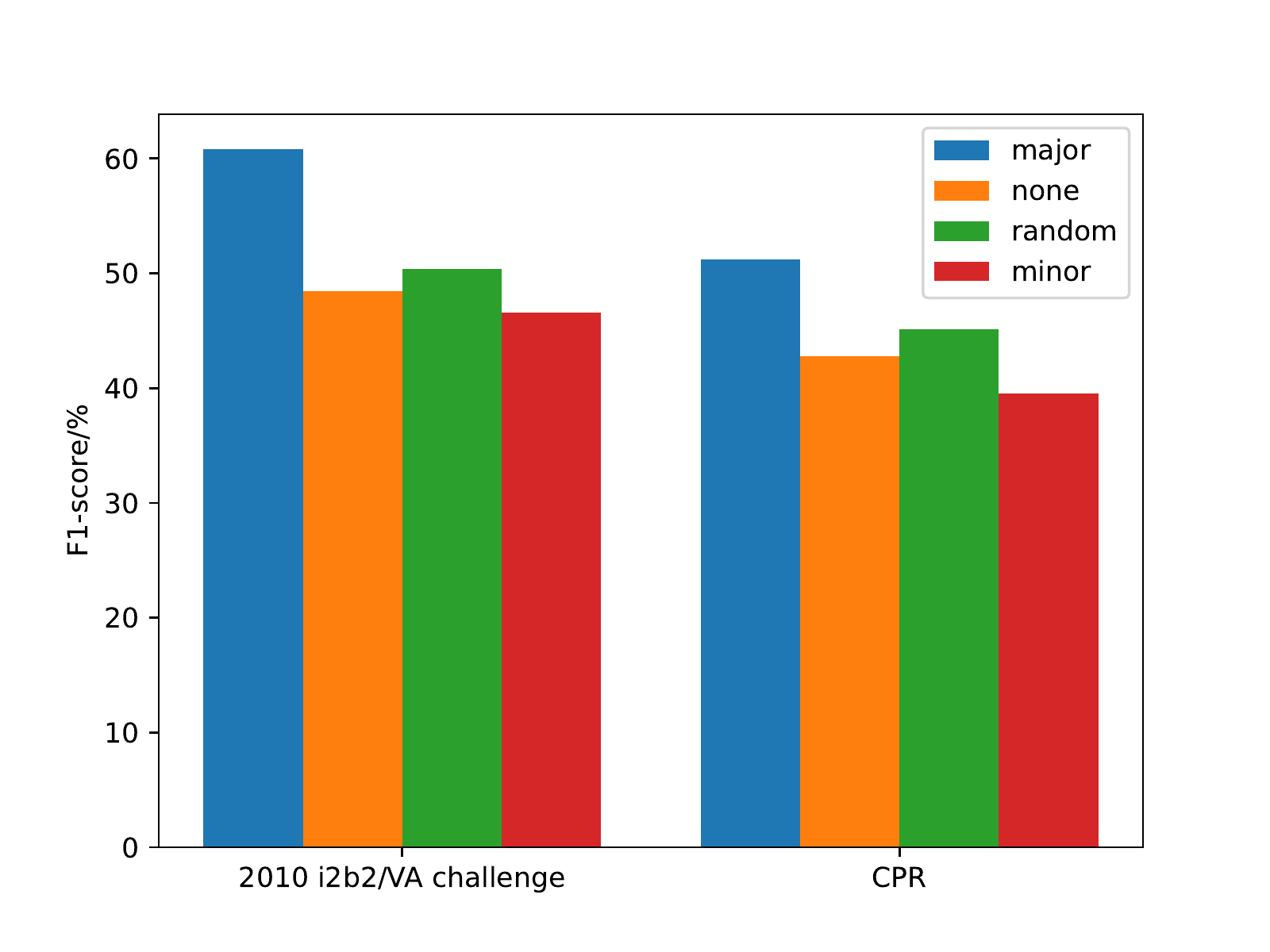}
\caption{Ablation results to analyze the effect of different classifier vectors.}
\label{ablation}
\end{figure}

We conducted experiments on the 2010 i2b2/VA challenge and CPR and set $\alpha=0.01$. As shown in Fig. (\ref{ablation}), \textit{major classifier vectors} is significantly better than \textit{random classifier vectors} and \textit{minor classifier vectors}. Among them, \textit{minor classifier vectors} achieves the worst performance, even inferior to \textit{none}, i.e., FedAvg without contrastive learning. This illustrates that different classifier vectors directly affect the effect of contrastive learning, and our similarity-based \textit{major classifier vectors} perform well.

Finally, we analyze the advantage of our algorithm compared to FedRS and FedLC. Both FedRS and FedLC are inspired by the study of imbalanced class distribution in centralized training \cite{longtail}\cite{bbn}. An underlying assumption of this problem is that the class distribution in the training set is imbalanced but that in the test set is balanced, so many class-balanced cross-entropy losses are designed. However, the global train and test distributions are agnostic and not always balanced in FL, which may be the reason for the poor performance of FedRS and FedLC. To this end, we conduct a small \textit{balanced} dataset by down-sampling 500 samples from CPR, 100 samples per class. And small \textit{unbalanced} dataset with 500 samples is also conducted following the label distribution in the original CPR. As shown in Table. \ref{balance}, It can be seen that FedRS and FedLC achieve similar performance to our algorithm under balanced global distribution. This shows that our algorithm is robust under various global distributions. 

\section{Conclusion}
In this paper, we find that the current FL algorithms, such as FedProx, MOON, and SCAFFOLD, have minor impacts on heterogeneous label distribution. One reason is that they do not take full advantage of all the clients' models, and another is that they are not designed specifically for the classifier. So, we propose FedCMC, which obtains ensemble major classifier vectors to correct the local training of individual clients. Our extensive experiments show that FedCMC outperforms the other state-of-the-art FL algorithms on various medical relation extraction datasets.

\bmhead{Acknowledgments}
This work was supported by the National Key Research and Development Program of China (2018YFC0830400), and the Shanghai Science and Technology Innovation Action Plan (20511102600).

\section*{Declarations}
\begin{itemize}
\item Funding Partial financial support was received from the National Key Research and Development Program of China (2018YFC0830400), and the Shanghai Science and Technology Innovation Action Plan (20511102600).
\item Conflict of interest/Competing interests
The authors have no relevant financial or non-financial interests to disclose.
\end{itemize}


\bibliography{sn-bibliography}


\end{document}